\pdfoutput=1

\documentclass[11pt]{article}

\usepackage{EMNLP2022}

\usepackage{times}
\usepackage{latexsym}
\usepackage{graphicx}
\usepackage{multirow}
\usepackage{multicol}

\usepackage[T1]{fontenc}

\usepackage[utf8]{inputenc}

\usepackage{microtype}

\usepackage{inconsolata}

\title{Clean Text and Full-Body Transformer: \\
Microsoft's Submission to the WMT22 Shared Task on\\ Sign Language Translation}

\author{*Subhadeep Dey,  Abhilash Pal, Cyrine Chaabani, *Oscar Koller \\
  Microsoft - Munich, Germany \\
  \texttt{\{subde,t-apal,t-cchaabani,oskoller\}@microsoft.com} \\}
  
\begin{document}
\maketitle
\def\thefootnote{*}\footnotetext{Equal Contribution}\def\thefootnote{\arabic{footnote}}
\begin{abstract}
This paper describes Microsoft's submission to the first shared task on sign language translation at WMT 2022, a public competition tackling sign language to spoken language translation for Swiss German sign language. The task is very challenging due to  data scarcity and an unprecedented vocabulary size of more than 20k words on the target side. Moreover, the data is taken from real broadcast news, includes native signing and covers scenarios of long videos. Motivated by recent advances in action recognition, we incorporate full body information by extracting features from a pre-trained I3D model and applying a standard transformer network. The accuracy of the system is further improved by applying careful data cleaning on the target text. We obtain BLEU scores of $\textbf{0.6}$ and 0.78 on the test and dev set respectively, which is the best score among the participants of the shared task. Also in the human evaluation the submission reaches the first place. The BLEU score is further improved to $\textbf{1.08}$ on the dev set by applying features extracted from a lip reading model. 
\end{abstract}

\section{Introduction}
Sign languages are natural visual languages that are used by deaf and hard of hearing individuals to communicate in everyday life. Sign languages are actively being researched. However, there is a huge imbalance in the field of natural language and speech processing between oral and signed languages. Since recently, one observes the emergence of a transition shifting sign language processing to be part of the NLP mainstream~\cite{yin_including_2021}. We embrace this development which manifests (among others) in the creation of the first shared task on sign language translation as part of WMT 2022~\cite{muller-etal-2022-findings}. 
It is great to have real-world sign language data~\cite{bragg_sign_2019,yin_including_2021} as the basis of this shared task, manifested in native signers content and an unprecedentedly large vocabulary. Nevertheless, this leads to a very challenging task with low performance numbers. 
When participating in the advances of sign language technologies it is worth recapping that deaf people have much at stake, both to gain and lose from applications that will be enabled here~\cite{bragg_fate_2021}. We aim to advance the field and the use-cases in a positive way and present our findings in this system paper.

In the remainder of this work we first present a brief view on the relevant literature in Section~\ref{sec:related_work}, then we present the employed data in Section~\ref{sec:data}. Subsequently, we describe our submission in Section~\ref{sec:submitted_system}, additional experiments in Section~\ref{sec:additional_experiments} and we end with a summary in Section~\ref{sec:summary}.

\section{Related Work}
\label{sec:related_work}

In this section, we present a limited overview of related work in sign language translation. We focus this review on the translation direction from sign language to spoken language and dismiss approaches that target the opposite direction, i.e. sign language production. 

Sign language translation started targeting written sign language gloss to spoken language text translations, hence no videos were involved. Related works were mainly based on phrase-based systems employing different sets of features~\cite{stein_hand_2007,stein_sign_2010,schmidt_using_2013}.
Then, neural machine translation revolutionized the field. The first research publications on neural sign language translation were based on LSTMs either with full image input~\cite{camgoz_neural_2018} or utilized human keypoint estimation~\cite{ko_neural_2019}.
Transformer models then replaced the recurrent architectures~\cite{camgoz_sign_2020a,yin_better_2020,yin_sign_2020}. These models perform a lot better, but suffer from a basic drawback that the input sequences must be limited to a maximum length. 
Previous work~\cite{camgoz_neural_2018,orbay_neural_2020} has identified the need for strong tokenizers to produce compact representations of the incoming sign language video footage. Hence, a considerable body of publications target creating tokenizer models that are often trained on sign language recognition data sets~\cite{koller_weakly_2020,koller_deep_2016,zhou_spatialtemporal_2022} or sign spotting data sets~\cite{albanie_bsl1k_2020,varol_read_2021,belissen_automatic_2019,pfister_largescale_2013}.

There are several data sets relevant for sign language translation. Some of the most frequently encountered are RWTH-PHOENIX-Weather 2014T~\cite{koller_continuous_2015,camgoz_neural_2018} and the CSL~\cite{huang_videobased_2018} (which could be also considered a recognition data set).
However, there are promising new data sets appearing: OpenASL~\cite{shi_opendomain_2022}, SP-10 dataset~\cite{yin_mlslt_2022} (covers mainly isolated translations) and How2Sign~\cite{duarte_how2sign_2021}.

\section{Data}
\label{sec:data}

To train our system, we used the training data provided by the shared task organizers. The data can be considered real-life-authentic as it stems from broadcast news using two different sources: FocusNews and SRF. FocusNews, henceforth FN, is an online TV channel covering deaf signers with videos of 5 minutes having variable sampling rates of either 25, 30 or 50 fps. SRF represents public Swiss TV with contents from daily news and weather forecast which are being interpreted by hearing interpreters. The videos are recorded with a sampling rate of 25 fps. All data, therefore, covers Swiss German sign language (DSGS).
Our feature extractors are pretrained on BSL-1k~\cite{albanie_bsl1k_2020} and AV-HuBERT~\cite{shi_learning_2022}. 
Additionally, we evaluate the effect of introducing a public sign language lexicon that covers isolated signs~\footnote{https://signsuisse.sgb-fss.ch/}, which we refer to as Lex. It provides main hand shape annotations, one or multiple (mostly one) examples of the sign and an example of how this sign is used in a continuous sentence. We choose a subset that overlaps in vocabulary with either FocusNews or SRF. 
As part of the competition, independent dev and test sets are provided, which consist of 420 and 488 utterances respectively.

Table~\ref{tab:train_data} shows the statistics of the training data. We see, that there is about 35 hours of training data in total. In raw form without any preprocessing the data is case sensitive, contains punctuation and digits. In this raw form the vocabulary amounts to close to 35k different words on the target side (which is written German). 22k words of these just occur a single time in the training data (singletons).
Through careful preprocessing as described in Section~\ref{sec:data_cleaning} we can shrink the vocabulary to about 22k words and the singletons to about 12k.

\begin{table}[t]
\begin{center}
        \begin{tabular}{ l  r  r  r|| r}
        \hline
             & \textbf{SRF} & \textbf{FN} & \textbf{Lex} & \textbf{Total} \\ \hline
            Videos & 29 & 197 & 1201 & 1427\\
            Hours & 15.6 & 19.1 & 0.9 & 35.6\\ \hline
            \multicolumn{5}{c}{Raw: no preprocessing} \\\hline
            Vocabulary & 18942 & 21490 & -- & 34783\\
            Singletons & 12433 & 13624 & -- & 22083 \\ \hline
            \multicolumn{5}{c}{Clean: careful preprocessing}\\\hline
            Vocabulary & 13029 & 14555 & 821 & 22840 \\
            Singletons & 7483 & 7923 & 591 & 12290\\    \hline
        \end{tabular}
        \caption{Data statistics on data used for training. SRF and FN refer to SRF broadcast and FocusNews data, while Lex stands for a public sign language lexicon. Singletons are words that only occur a single time during training. }
        \label{tab:train_data} 
     \end{center}
\end{table}

\section{Submitted System}
\label{sec:submitted_system}

Sign languages convey information through the use of manual parameters (hand shape, orientation, location and movement) and non-manual parameters (lips, eyes, head, upper body).
To capture most information from the signs, we opt for an RGB-based approach, neglecting the tracked skeleton features by the shared task organizers. For the submitted system we rely on a pre-trained tokenizer for feature extraction and train a sequence-to-sequence model to produce sequences of whole words (no byte pair encoding).
We further pre-process the sentences (ground truths of the videos) to clean it.
This step is crucial to push the model to focus more on semantics of the data. 
Finally, in order to adhere to the expected output format for the submission, we convert the text back to display format using Microsoft's speech service. This applies inverse text normalization, capitalization and punctuation to the output text to make it more readable. The details of various components of the system are described in the next subsections. 

\subsection{Features}

We use a pre-trained I3D~\cite{carreira_quo_2017a} model, based on inflated inceptions with 3D convolutional neural networks, to extract features for our task.
The model~\cite{varol_read_2021} was pre-trained to take consecutive video frames as input and predict over 1k sign classes. It was trained on BSL-1K~\cite{albanie_bsl1k_2020} consisting of about 700k spotted sign instances from the British broadcast news.
The features are extracted with a context window of 64 frames and a temporal stride of 8. We use the model as a feature extractor, recovering embeddings before the final classification layer (layer: mixed\textunderscore 5c), yielding a sequence of 1024 dimensional vector, extracted for each video.
Our input data is required to match the training conditions of the network, hence we apply gray background padding (adding 20\% padding left and right, 7.5\% up and down) and rescale the videos to 224 × 224 resolution. The front end features are subsequently fed to a sequence model.

\subsection{Sequence model}

A standard transformer network is trained to predict text sequences. We apply word-based units as the output instead of byte pair encoding. It seems that full words help reduce ambiguity in a data constrained scenario. The model is trained with the fairseq~\cite{ott_fairseq_2019} toolkit. We apply 3 transformer layers at the encoder side and 2 layers at the decoder with 1024 hidden feed-forward dimension and 22k output units. The model is trained with Adam optimizer for 2k epochs with a learning rate of 1e-3. We found that a beam size of 1 works well on the dev set during decoding. 

\subsection{Data Cleaning}
\label{sec:data_cleaning}

To reduce ambiguity and noise in the target text, we first applied manual cleaning by removing
foreign (French and English) sentences. We then proceeded to
removing sentences that start with a hashtag sign, as this seems to indicate inaccurate annotation.
Further, we removed status messages that were added by the subtitling agency (such as ``1:1-Untertitelung.'',
``Livepassagen können Fehler enthalten.'',
``Mit Live-Untertiteln von SWISS TXT'')
and patterns enclosed by an asterisk which indicate sounds occuring in the show (e.g. ``* Beschwingte Blasmusik *''). 
As a next step, we expanded abbreviations like ``Mrd.'' to ``Milliarden'' and applied text normalization to remove
punctuation and special characters, lower case of the text and expand numbers and dates. As can be shown in Table~\ref{tab:train_data}, this plays a major role in reducing the total vocabulary.

\subsection{Evaluation metrics}

A common evaluation metric for machine translation is BLEU~\cite{papineni_bleu_2002}. However, the difficulty of the given task and the inherently low performance of the submitted systems cause a bias in the automated evaluation. Spoken languages like Swiss German, which is the target output space for the translation in this challenge, follow a statistical pattern where stop words or function words constitute the classes of words that occur most frequently. This explains one of the observations that we made in regard to the generated system output. In fact, the models which were producing more stop words achieved the highest BLEU scores.
For example, the model we submitted achieved a BLEU score of 0.77 on the dev set, while an earlier, clearly worse, checkpoint achieved 0.91. Looking at the stop words, the submitted model output counts 2125 stop words, while the earlier checkpoint counts 2237 of such words. After our stop word removal the submitted reduced BLEU score is 0.78, while the earlier model achieves only 0.66.
Hence, for model selection, we propose to use `reduced BLEU' with an additional step:  we first apply a blacklist to remove stop and function words. Using this approach, the model selection process based on the reduced BLEU metric turned to be much more reliable and more reflective of actual performance. In this work, we report both reduced BLEU and standard BLEU for all results on the dev set. Only standard BLEU is reported by the automatic evaluation through the shared task on the test set. The list of employed stop words can be found in the appendix.

\subsection{Results}

\begin{table}[t]
\begin{center}
        \begin{tabular}{ r  r r  r }
        \hline
             Data cleaning & Dev (RedB) & Dev  & Test   \\ \hline
              no &  0.49 & 0.70  & 0.4 \\
              yes &  $\textbf{0.78}$ & \textbf{0.77}  & $\textbf{0.6}$ \\\hline
        \end{tabular}
        \caption{The Table shows the effect of data cleaning. Performance of translation systems trained on SRF and FocusNews evaluated on the WMT 2022 dev and test data is provided. We report reduced and standard BLEU score on the Dev set and only standard BLEU on the Test set. RedB stands for the reduced BLEU measure.}
        \label{table:results_submitted_system}
        \end{center}
        \end{table}
The results of the submitted systems are presented in Table~\ref{table:results_submitted_system}, which allows to compare the effect of applying data cleaning.
In terms of reduced BLEU, data cleaning improves the performance from 0.49 to 0.78 on the dev data. The test data shows similarly an improvement from 0.4 BLEU to 0.6.
 Based on the preliminary automatic BLEU scoring this result was the best among the participants of the shared task. However, it has to be noted that the final evaluation will be based on a human eval that has not yet been completed at the time of paper submission.
It can be concluded that careful data preparation is fundamental for this data. Nevertheless, the shared task proves to be very challenging and overall, we observe rather low performance compared to published results on benchmark data sets like RWTH-PHOENIX 2014T~\cite{camgoz_neural_2018}. This further proves the large amount of variability in the task, which is amplified due to the presence of high numbers of singleton words.

\section{Additional Experiments}
\label{sec:additional_experiments}

We perform additional experiments to assess the impact of having dedicated mouth features as well as a lexicon data set on the model performance. The mouth carries important semantic information in sign languages. In the literature, exploiting lip information has shown to be fundamental for increased performance in sign language recognition and translation~\cite{koller15:mouth,shi_opendomain_2022}. 

\subsection{Mouth features} 
To extract features from the lip area, we employ a pre-trained AV-HuBERT model~\cite{shi_learning_2022} that has been trained on English data. AV-HuBERT is trained to learn a robust audio-visual representation in a self-supervised fashion. The success of the model is evident by the performance on an audio-visual speech recognition task. It has proven to be useful for sign-language task as well~\cite{shi_opendomain_2022}.
For us, the first step is to obtain mouth patches from the video frames. Hence, we rely on the dlib utility provided by the AV-HuBert authors for obtaining facial key point extraction.  The face patch is cropped and re-scaled to match the input size (96 x 96) of the model. We use the AV-HuBERT model to extract 768 dimensional embedding (output of the ResNet layer) to obtain a feature vector per frame.

\subsection{Comparison to RWTH-PHOENIX 2014 T}

To underline the difficulty of the given shared task, we compare our employed pipeline on a standard benchmark data set for sign language translation, namely Phoenix 2014T~\cite{camgoz_neural_2018}. We noticed a small difference between the original PHOENIX 2014T corpus as referenced and shared in of~\cite{camgoz_neural_2018} and the publicly available embeddings and experiments of~\cite{camgoz_sign_2020a}. In the latter full stops mark the end of each utterance, while in the original version this is not the case. The effect is small, but for the sake of completeness, we show it here. Furthermore, considering the statistics of PHOENIX 2014T and this WMT 2022 shared task, the difference becomes apparent. Table~\ref{tab:data_comparison} shows the key statistics for the two tasks side by side. We can see that the WMT 2022 task has a nearly 8 times larger vocabulary, with 11 times more singletons that occur only once in training. However, it has not even 4 times more video material.

\begin{table}[t]
\begin{center}
        \begin{tabular}{ l  r  r  r}
        \hline
             & WMT & PHOENIX & \multirow{2}{*}{Factor} \\
             &2022& 2014T&\\\hline
            Hours & 35.6 & 9.2 & 3.9\\
            Vocabulary & 22840 &2887& 7.9 \\
            Singletons & 12290& 1077& 11.4 \\\hline
            
        \end{tabular}
        \caption{Comparison between the training data for the WMT 2022 shared task and PHOENIX 2014T.}
        \label{tab:data_comparison} 
 \end{center}
\end{table}

\subsection{Results}

\begin{table}[t]
\begin{center}
        \begin{tabular}{ l  l r r }
        \hline
            &\multirow{2}{*}{Training data} &  \multicolumn{
            2}{c}{Features}\\
            & & Full body  & Mouth  \\ \hline
           \parbox[t]{2mm}{\multirow{2}{*}{\rotatebox[origin=c]{90}{RedB}}}  & SRF + FN &  0.78   & 0.95 \\ 
             &SRF + FN + Lex &  0.54  & 1.08 \\
            \hline
          \parbox[t]{2mm}{\multirow{2}{*}{\rotatebox[origin=c]{90}{Stand.}}}   & SRF + FN &  0.77   & 1.15 \\ 
            & SRF + FN + Lex &  0.68  & 1.27 \\
            \hline
        \end{tabular}
        \caption{The effect of adding lexicon data to systems trained with full body features (I3D) and mouth features (AV-HuBERT). Configurations are evaluated on the WMT 2022 Dev dataset using the reduced BLEU (RedB) and standard BLEU (Stand.) score as metric.}
        \label{table:results_additional_experiment}
        \end{center}
        \end{table}

Table~\ref{table:results_additional_experiment} shows the results of applying AV-HuBERT features as input to the sequence model. It can be observed that the model trained with AV-HuBERT features performs better than the I3D model. In fact, it achieves a 0.95 reduced BLEU score, while the full body I3D features reach only 0.78. On visual inspection, we found that the model trained with AV-HuBERT is able to predict infrequent words but fails on simple words such as  "auf wiedersehen". Therefore, we assess the effect of adding a lexical data set (`Lex') to boost representations of those simple words.
Table~\ref{table:results_additional_experiment} shows that the addition of lexical data further improves the model performance which reaches 1.08 reduced BLEU. We believe that this is likely due to the matching lip movement patterns in the lexical training dataset and dev dataset. Unfortunately, due to time constraints, we were not able to submit this model.

Table~\ref{table:results_phoenix} shows results on PHOENIX 2014T. We can see that our submitted pipeline matches the performance of~\cite{camgoz_sign_2020a} (19.80/20.24 BLEU on the dev/test sets compare to 20.69/20.17). However our employed embeddings, which were not trained on PHOENIX 2014T, do not generalize well to PHOENIX 2014T and are hence significantly outperformed by the ones employed in~\cite{camgoz_sign_2020a}. The experiments show that the WMT 2022 shared task is significantly more challenging than PHOENIX 2014T. We also see that the addition of full stops at the end of each utterances in PHOENIX 2014T amounts to a difference in BLEU of about 1\% relative (14.22/13.22 BLEU with full stops opposed to 14.06/13.13 without full stops).

\begin{table*}[t]
\begin{center}
        \begin{tabular}{  l r r r r r }
        \hline
            \multirow{2}{*}{Approach} &  Decoding &  \multicolumn{2}{c}{Fullstops} &  \multicolumn{2}{c}{PHOENIX 2014T}\\
            &Beam&No&Yes& Dev  & Test  \\ \hline
             \cite{camgoz_sign_2020a} &  10 & & x & 20.69 & 20.17 \\ 
              Our pipeline with embeddings from \cite{camgoz_sign_2020a} &  5 & & x & 19.80 & 20.24 \\
              Our pipeline with WMT full body embeddings & 5 & x &  & 14.06 & 13.13 \\
              Our pipeline with WMT full body embeddings & 5 & & x & 14.22 & 13.22 \\
            \hline
        \end{tabular}
        \caption{Showing results on PHOENIX 2014T, a benchmark data set. The system which matches our submission to the shared task can be found on the last line.}
        \label{table:results_phoenix}
        \end{center}
        \end{table*}

\section{Summary}
\label{sec:summary}
In this paper, full body information is applied successfully for a challenging sign language task as part of the WMT 2022 competition. As such, we employed a pre-trained I3D model to extract an embedding for a sequence of frames of the video. The features are further fed as input to a standard transformer network. We obtain reasonable performance of 0.4 in terms of BLEU score on the test set. The model is further enhanced by applying careful cleaning to the text output. We obtain the result of 0.6 BLEU score on the official test data. Based on the automatic BLEU scoring this result was the best among the 7 participants of the shared task, but also in the human evaluation our submission reaches the first place.
With additional experiments, we validate the usefulness of a pre-trained lip reading model for this task and the addition of a lexical data set. This improves the results to 1.08 reduced BLEU on the dev set. 

\section*{Limitations}

One major limitation to our work resides within the data set used for training our model. In fact, the signing interpreter is usually not a native signer and often seems to be heavily influenced by the source language, a.k.a the spoken language. As stated previously, we used a signing interpreter for SRF data. 
Another issue we have identified lies in the limited domain of the data, as it is constrained to Broadcast news. The trained model may therefore be too specialized to generalize well beyond this area. Furthermore, due to the small number of individuals present in the data set, it remains unclear if and how much ethnicity bias is introduced to the model. Our team did not proceed with any experiments to identify and measure this. However, we do believe that it is crucial to further analyze possible biases in the future. One evaluation metric that we did take into consideration for the model performance is the BLEU score. The experiments consistently returned extremely low values which reflects a poor accuracy. One thing that remains unclear to us is how significant small BLEU differences are for human perception and subjective evaluation.

\bibliography{koller}
\bibliographystyle{acl_natbib}
\appendix
\onecolumn
\begin{multicols}{5}
[
\section{List of Stop Words for Reduced BLEU Estimation}
]
ab\\
als\\
als\\
also\\
am\\
am\\
an\\
an\\
andere\\
auf\\
aus\\
beim\\
bin\\
bist\\
da\\
darauf\\
das\\
dass\\
davon\\
dazu\\
dem\\
den\\
denen\\
der\\
des\\
des\\
deshalb\\
dessen\\
die\\
dies\\
diese\\
diesen\\
dieser\\
dieses\\
doch\\
dort\\
ein\\
eine\\
einem\\
einen\\
einen\\
einer\\
eines\\
eines\\
er\\
es\\
es\\
für\\
gar\\
gegen\\
geht's\\
genau\\
gibt\\
habe\\
haben\\
habt\\
hast\\
hast\\
hat\\
hat\\
hatte\\
hätte\\
hatten\\
hätten\\
her\\
hin\\
ihm\\
ihre\\
ihre\\
im\\
in\\
ins\\
ist\\
könne\\
könnte\\
könnten\\
man\\
mehr\\
mit\\
noch\\
nun\\
ob\\
oder\\
quasi\\
schon\\
sehr\\
sei\\
seid\\
seien\\
sein\\
seit\\
sich\\
sie\\
sie\\
sind\\
so\\
solchen\\
soll\\
somit\\
sowie\\
sowohl\\
statt\\
über\\
um\\
und\\
vom\\
von\\
vor\\
war\\
war\\
wäre\\
war's\\
wars\\
warst\\
wart\\
wegen\\
weiteren\\
weiterhin\\
wem\\
wen\\
wenn\\
werde\\
werden\\
werdet\\
weshalb\\
wie\\
will\\
wir\\
wird\\
wirst\\
wo\\
wohl\\
wolle\\
wollte\\
wollten\\
worauf\\
wurde\\
würde\\
würden\\
zu\\
zudem\\
zum\\
zur\\
zur\\
zur\\
zwar\\

\end{multicols}

\end{document}